\documentclass[10pt,conference]{IEEEtran}
\usepackage{graphicx}
\usepackage{tabularx}
\usepackage{booktabs}
\usepackage{amsmath} 
\usepackage{hyperref}
\usepackage{xcolor}
\usepackage{balance}
\usepackage{algorithm}
\usepackage{enumitem}
\usepackage[noend]{algpseudocode}
\usepackage{adjustbox}
\usepackage{comment}

\title{MSLEF: Multi-Segment LLM Ensemble Finetuning in Recruitment}

\author{
\IEEEauthorblockN{{\large Omar Walid\IEEEauthorrefmark{1},
Mohamed T. Younes\IEEEauthorrefmark{1},
Khaled Shaban\IEEEauthorrefmark{2},
Mai Hassan\IEEEauthorrefmark{1},
Ali Hamdi \IEEEauthorrefmark{1}}}

\IEEEauthorblockA{\IEEEauthorrefmark{1}\large\textit{Dept. of Computer Science, MSA University, Giza, Egypt} \\
\{ omar.walid2, mohamed.tarek61,maisalem, ahamdi \}@msa.edu.eg}

\IEEEauthorblockA{\IEEEauthorrefmark{2}\large\textit{Dept. of Computer Science, Qatar University, Doha, Qatar} \\
khaled.shaban@qu.edu.qa}
}

\begin{document}

\maketitle

\begin{abstract}
This paper presents MSLEF, a multi-segment ensemble framework that employs LLM fine-tuning to enhance resume parsing in recruitment automation. It integrates fine-tuned Large Language Models (LLMs) using weighted voting, with each model specializing in a specific resume segment to boost accuracy. Building on MLAR \cite{mlar2025}, MSLEF introduces a segment-aware architecture that leverages field-specific weighting tailored to each resume part, effectively overcoming the limitations of single-model systems by adapting to diverse formats and structures. The framework incorporates Gemini-2.5-Flash LLM as a high-level aggregator for complex sections and utilizes Gemma 9B, LLaMA 3.1 8B, and Phi-4 14B. MSLEF achieves significant improvements in Exact Match (EM), F1 score, BLEU, ROUGE, and Recruitment Similarity (RS) metrics, outperforming the best single model by up to +7\% in RS. Its segment-aware design enhances generalization across varied resume layouts, making it highly adaptable to real-world hiring scenarios while ensuring precise and reliable candidate representation.

\end{abstract}

\section{Introduction}
Recruitment automation has transformed hiring by enabling efficient processing of numerous applications, with resume parsing extracting structured data like experience, education, skills, and contact information. Challenges from varied layouts (e.g., plain text to multi-column designs) and complex language lower accuracy and delay candidate matching, especially as some roles attract thousands of resumes \cite{li-etal-2025-karrierewege}, making manual screening impractical. \par
The impact of poor resume parsing affects employers and applicants alike. Large companies may overlook top talent, leading to longer vacancies, higher costs, and reduced productivity. Strong candidates can be rejected due to confusing formats or wording, exacerbating hiring challenges as skilled worker demand grows. Past methods like rule-based extraction and NER \cite{lample2016named}, and conditional random fields \cite{lafferty2001conditional}, suit clean text but struggle with diverse layouts. Improved systems are essential, as poor parsing wastes talent and raises costs \cite{herbold-etal-2025-impact}. Biased systems harm fairness by ignoring varied layouts or styles \cite{iso-etal-2025-evaluating} and pose legal risks. Internationally, recruitment AI is increasingly classified as ``high-risk.'' Under the EU AI Act (2024), any AI system used for employment, recruitment, or worker management is explicitly designated as a high-risk system requiring strict obligations, including risk management, transparency, human oversight, and robust documentation~\cite{eu_ai_act_2024}. \par

\begin{figure*}[t]
  \centering
  \includegraphics[width=0.75\textwidth]{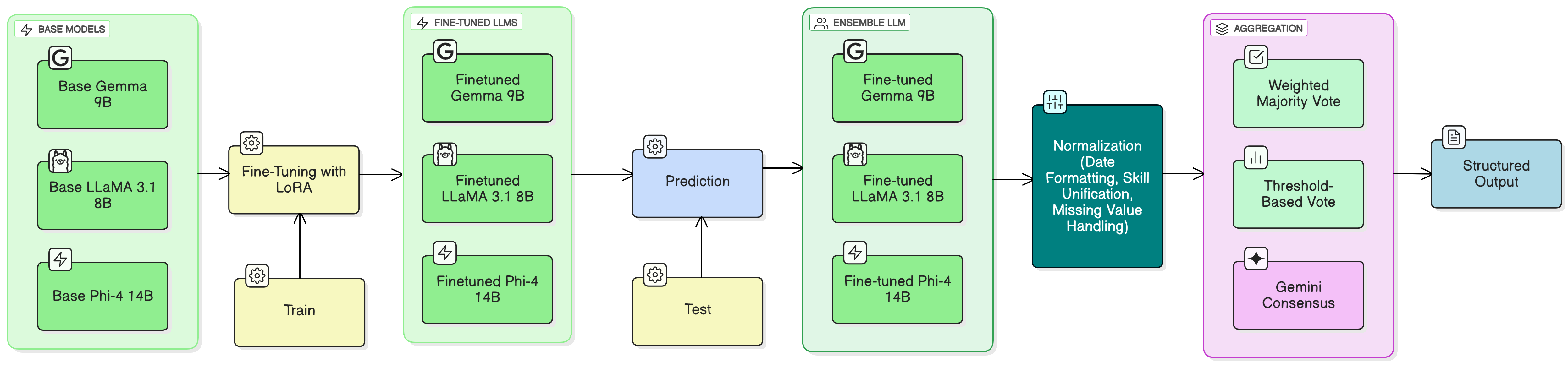}
  \caption{MSLEF workflow for resume parsing and job matching. Multiple fine-tuned LLMs analyse distinct resume segments and are aggregated via weighted voting with Gemini-LLM for complex hierarchies.}
  \label{fig:Sys_flowchart}
\end{figure*}

To solve these issues, we propose MSLEF, a framework that uses multi-segment LLM ensemble fine-tuning to improve parsing. Each part of the resume, the header, skills, timeline, and free text, is handled by a specific fine-tuned model. MSLEF combines the output with a field-weighted vote. Gemini LLM resolves conflicts in complex sections. It uses Gemma 9B, LLaMA 3.1 8B, and Phi-4 14B as specialised base models. MSLEF scores 92.6\% F1, compared to 90.6\% for the best single model, Phi-4 14B. This gain cuts residual error by about 21\%. Exact match and recruitment-based metrics also improve. Error rates on unusual layouts drop by over one-third. The MSLEF design uses the strengths of the model per segment for better accuracy~\cite{abburi-etal-2023-simple}. This method helps to handle various formats and complex wording in resumes~\cite{vaishampayan-etal-2025-human}.\par

Our contributions are:
\begin{itemize}
    \item A segment-aware, model-agnostic merging architecture that integrates heterogeneous LLMs for flexible resume parsing through \textbf{multi-segment ensemble finetuning}.
    \item A field-specific weighting strategy for consensus voting that adjusts the influence of each model according to its segment proficiency.
    \item A hybrid evaluation suite that combines standard NLP metrics with recruitment-centric criteria, together with an open benchmarking dataset and reproducible test harness.
\end{itemize}
These contributions represent a significant advancement in the field of recruitment automation, offering a more robust and adaptable solution for resume parsing that can handle the complexities of modern job applications \cite{chowdhury-etal-2025-natural}. The practical implications of this work extend to improving the efficiency of hiring and the experience of candidates in diverse industries. \par
The remainder of this paper is organised as follows. Section \ref{RelatedWork} surveys related work on LLMs in recruitment and ensemble NLP. Section \ref{Methodology} outlines our methodology, including task formulation and synthetic dataset creation. Section \ref{MSLEF Framwork} presents the MSLEF Framework, detailing the fine-tuning of the LLMs, ensemble architecture, and evaluation metrics. Section \ref{ExperimentalDesign} describes the experimental design, Section \ref{Results and Discussion} reports the results and discussion, and Section \ref{Conclusion} concludes with directions for future research.\par

\section{Related Work} \label{RelatedWork}
Large Language Models (LLMs) are increasingly used in recruitment tools for resume parsing and job matching. Vagale et al.~\cite{vagale2024prospectcv} proposed ProspectCV, an LLM platform that compares resumes to job posts. It achieved a high level of precision in skill extraction on curated data, but struggled with real-world layout diversity. Abdollahnejad et al.~\cite{abdollahnejad2021bert} applied BERT for person–job fit, reporting high accuracy on structured candidate profiles yet showing limitations on mixed or unstructured resumes. Our earlier system, MLAR \cite{mlar2025}, achieved an F1 score of 0.81 and demonstrated the value of robotic process automation in hiring. Chowdhury et al.~\cite{chowdhury-etal-2025-natural} surveyed HR automation trends, noting that LLMs improve parsing accuracy compared to traditional methods and emphasizing the need for ensemble strategies. These studies collectively highlight both the promise and limitations of current approaches, motivating MSLEF’s segment-aware ensemble design for diversified resume data. \par
Ensemble methods help NLP systems perform better on varied datasets. They support generalization across formats and tasks. Yu et al.~\cite{yu-etal-2025-confit} achieved 13.8\% recall and 17.5\% nDCG gains in resume-job matching using hypothetical resume embeddings and runner-up hard-negative mining with transformer encoders. They noted overfitting risks in iterative training. We address this with field-specific optimization. Schaudt et al.~\cite{schaudt2023combining} combined RPA and AI, improving document classification by 9\%. However, they did not use field-specific weights like MSLEF. Our method adjusts model votes based on strength per field, such as skills or education. These works prove that ensembles help in NLP. But none focus on resume parsing with multi-model voting. \par
Other studies in ensemble methods and structured extraction guide MSLEF’s design. Romero et al.~\cite{romero-etal-2025-medication} used stacked and voted LLM ensembles for medication and entity extraction. They reached high precision on clinical texts. Their setup mirrors our use of segment-specific models for resumes. Jiang et al.~\cite{jiang-etal-2023-llm} proposed LLM-Blender, which blends outputs with ranking and generation. This idea supports our weighted voting by field strength. Tekin et al.~\cite{tekin-etal-2024-llm} created LLM-TOPLA, which improves accuracy through diverse LLMs. MSLEF follows this by mixing different models across resume parts. \par
Furthermore, Layout-aware models like DocLLM~\cite{wang-etal-2024-docllm} improve document understanding using layout data. These models help with visually rich documents. Their success shows that layout matters in parsing. This supports MSLEF’s focus on structured resume layouts. Iso et al.~\cite{iso-etal-2025-evaluating} studied bias in LLMs for job-resume matching. They stressed fairness in automated hiring. The importance of fairness is underscored by real-world failures, such as Amazon’s abandoned AI tool that was biased against women~\cite{dastin_2018}, contravening standards like ILO Convention No. 111~\cite{ilo_1958}. MSLEF addresses this by using diverse models to lower bias and raise accuracy. \par

\section{Methodology} \label{Methodology}
\subsection{Task Definition}
The task of resume parsing in recruitment automation involves extracting structured information, such as name, email, phone, skills, experience, education, and department from unstructured or semi-structured resume documents. Because resumes contain highly sensitive personal data, any automated parsing engages data-protection regimes. Under the GDPR, core principles apply: purpose limitation (Art. 5(1)(b)), data minimization (Art. 5(1)(c)), and restrictions on fully automated decision-making affecting individuals (Art. 22). These provisions mandate transparency, candidate rights of explanation and contestation, and human-in-the-loop safeguards. MSLEF’s architecture, particularly its emphasis on transparency of parsing steps and Gemini-based consensus, can be positioned as compatible with these requirements~\cite{gdpr_2016}. 

Formally, the task is to map $d$ to a structured output $s = \{f_1, f_2, \ldots, f_k\}$, where each $f_i$ represents a specific field (e.g., $f_1 = \text{name}$, $f_2 = \text{skills}$, etc.). This requires handling diverse formats and extracting semantically rich data, such as multi-part education or hierarchical experience, where single-model LLMs often falter due to limited generalization and inconsistent outputs. To address this, we propose an ensemble approach integrating fine-tuned LLMs Gemma 2 9B, LLaMA3.1 8B, and Phi-4 14B specialized for different parsing aspects. The ensemble uses a weighted multi-voting mechanism, applying majority votes for simple fields and a Gemini LLM hybrid strategy for complex fields like education and experience, enhancing accuracy and robustness across diverse resume types.

\subsection{Synthetic Dataset Creation}
The dataset is a hybrid corpus of 2,400 real-world resumes from the Kaggle Resume Dataset \cite{kaggle2021resume} (PDFs across 24 job titles, English) and 1,000 synthetic resumes. Real resumes are parsed into a standardized JSON format using DeepSeek \cite{deepseekai2025}, extracting fields: name, email, skills, education, experience, and department.\par
Synthetic resumes are generated using Python scripts and data programming techniques inspired by Ratner et al. \cite{ratner2017snorkel} to enhance diversity and target edge cases, then converted to JSON. The merged dataset is normalized by standardizing dates (YYYY-MM-DD), unifying skill terms, and filling missing values with placeholders. It is split into training (80\%, 2,720 resumes), validation (10\%, 340 resumes), and testing (10\%, 340 resumes) sets, ensuring balanced representation across formats and professions. The process progresses from parsing real resumes, generating synthetic ones, merging them, normalizing data, and splitting for training, validation, and testing.\par
This hybrid dataset supports fine-tuning of Gemma 9B, LLaMA3.1 8B, and Phi-4 14B using Low-Rank Adaptation (LoRA), as shown in the final stage of Figure \ref{fig:Sys_flowchart}, producing "Fine-Tuned Gemma 9B," "Fine-Tuned LLaMA3.1 8B," and "Fine-Tuned Phi-4 14B" for the MSLEF pipeline. The process enables the ensemble to effectively handle diverse resume structures and semantic nuances.

\section{MSLEF Framework} \label{MSLEF Framwork}
\subsection{Fine-Tuning of the LLMs}
To adapt the LLMs for the resume parsing task, we separately fine-tuned Gemma 2 9B, LLaMA3.1 8B, and Phi-4 14B on our hybrid dataset using Low-Rank Adaptation (LoRA), a parameter-efficient fine-tuning method. The fine-tuning process optimized each model for specific aspects of resume parsing, leveraging cross-entropy loss for simple fields (e.g., name, email) and sequence-to-sequence loss for complex fields (e.g., experience, education). Key hyperparameters included a LoRA rank of 16, a learning rate of 2e-4, and training for 3 epochs, ensuring efficient adaptation to the recruitment domain. These settings were selected based on validation performance, balancing accuracy and computational efficiency. This fine-tuning enables each LLM to contribute specialized strengths to the ensemble, as described in the next subsection. \par

\subsection{Ensemble Architecture}
The \textbf{MSLEF} framework integrates three fine-tuned LLMs Gemma 9B, LLaMA 3.1 8B, and Phi-4 14B chosen for their complementary strengths in resume parsing. Gemma delivers computational efficiency for large batches, LLaMA excels at contextual nuance (skills and free-form experience), and Phi-4 yields the most reliable structured outputs (education, dates). The pipeline, therefore, proceeds in three stages: output generation, normalization, and aggregation.

\paragraph*{Output generation.}
Each model \(m\!\in\!\mathcal{M}\) is prompted with field-specific instructions (e.g., “\textit{Extract the candidate’s name}”), producing
\begin{equation}
P_m \;=\; \text{extractFields}_m(r) \tag{1}
\end{equation}
a flat JSON object that lists candidate values for \emph{name}, \emph{email}, \emph{phone}, \emph{skills}, \emph{experience}, \emph{education}, and \emph{department}.

\paragraph*{Normalization.}
Formatting and vocabulary differences are reconciled against the full field set
\begin{equation}
\mathcal{F}_{\mathrm{all}}\;
 =\;\{\text{name},\text{email},\text{phone},\text{skills},\text{experience},\text{education},\text{dep.}\}.    
\end{equation}

The canonical form is obtained via
\begin{equation}
N_m \;=\; \text{NormalizeFields}\!\bigl(P_m,\mathcal{F}_{\mathrm{all}}\bigr) \tag{2}
\end{equation}
which standardizes dates (YYYY-MM-DD), folds skill synonyms into an ontology, and fills missing values with \texttt{N/A}.

\paragraph*{Aggregation.}
A weight vector \(\omega=\{\text{Phi}:3,\;\text{Gemma}:2,\;\text{LLaMA}:1\}\), tuned on a validation set of diverse resumes to reflect each model’s accuracy on specific field types, guides consensus.

\noindent
\emph{Scalar fields} (\emph{name}, \emph{email}, \emph{phone}, \emph{department}) are resolved through a weighted majority:
\begin{equation}
S[f] \;=\; \text{WeightedMajorityVote}(V,W), \tag{3}
\end{equation}
where \(V\) is the list of candidate values and \(W\) the corresponding model weights collected for field \(f\).

\noindent
\emph{Skills} are consolidated by first pooling every model’s list into the multiset represented by \(V\) and then retaining only those items whose cumulative support exceeds half the maximum attainable weight:
\begin{equation}
S[\text{skills}] \;=\;
\text{WeightedThresholdVote}\!\bigl(V,W,\tfrac12\!\sum_{m}\omega[m]\bigr), \tag{4}
\end{equation}
ensuring that no single low-weight model can dominate the final list.

\noindent
\emph{Experience} and \emph{education} are nested structures.  
If more than one model proposes conflicting sub-trees, they are fused through contextual reasoning over the original resume:
\begin{equation}
\begin{aligned}
   S[f] \;=\; \text{GeminiConsensus}(V,W), \\ 
   f \in\{\text{experience},\text{education}\}, 
\end{aligned} \tag{5}
\end{equation}
otherwise, the unique candidate passes through unchanged.  
Normalisation placeholders propagate when \emph{all} models exclude a field, providing an explicit signal to downstream components. While effective for standard resumes. \par

This mathematically grounded procedure yields the robust, field-specific consensus stored in the structured dictionary \(S\) returned. \par

\subsection{Evaluation Metrics}
We evaluate using Exact Match (EM), F1 Score, BLEU Score, ROUGE Score, and Recruitment Similarity (RS). Recruitment Similarity is our composite, HR-focused metric that weights key fields by their practical importance to recruiters:
\begin{equation}
\text{RS} = 0.35\,S_{\text{skills}} \;+\; 0.15\,S_{\text{email}} \;+\; 0.15\,S_{\text{phone}}
  \;+\; \sum_i w_i\,S_i ,
\tag{1}    
\end{equation}
where \(S_{\text{field}}\) denotes the similarity score (e.g.\ Levenshtein or cosine) for that field and the remaining weights \(w_i\) sum to 0.35.  
By emphasising \emph{skills} and reliable \emph{contact details}, RS directly reflects how hiring pipelines prioritise candidate information, while still allowing custom weighting for additional fields such as \emph{certifications} or \emph{languages}.

\medskip

\section{Experimental Design} \label{ExperimentalDesign}
\begin{itemize}[leftmargin=*]
\item \textbf{Hybrid dataset.}  
 2400 real‐world Kaggle resumes \emph{+} 1000 synthetically generated resumes
 form an 80 / 10 / 10 train–validation–test split, covering 24 professions
 and diverse layouts.

\item \textbf{Model fine-tuning.}  
 Gemma 9B, LLaMA 3.1 8B, and Phi-4 14B are fine-tuned with
 LoRA (rank = 16, 3 epochs, 2e-4 learning rate) on the training split.

\item \textbf{Three-stage pipeline.}  
 \emph{(i) Output generation} by each LLM  
 \emph{(ii) Normalisation} of dates, skills, and missing values  
 \emph{(iii) Aggregation} via weighted voting and Gemini-based
 consensus for complex fields.

\item \textbf{Weight calibration.}  
 Validation data determines the voting vector
 $\omega=\{\text{Phi}\!:\!3,\ \text{Gemma}\!:\!2,\ \text{LLaMA}\!:\!1\}$,
 giving more influence to models that excel on structured fields.

\item \textbf{Baselines and metrics.}  
 Each fine-tuned model is evaluated individually and then compared with
 the MSLEF ensemble using EM, F1, BLEU, ROUGE, and the
 HR-centric Recruitment Similarity (RS) score.
\end{itemize}

The MSLEF framework enhances resume parsing for recruitment automation by integrating fine-tuned Large Language Models (LLMs)—Gemma 2 9B, LLaMA 3.1 8B, and Phi-4 14B—through a three-stage pipeline: output generation, normalization, and aggregation. It processes preprocessed resume text into structured JSON output (name, email, phone, skills, experience, education), overcoming single-model limitations with weighted multi-voting and GeminiLLM synthesis for complex fields.

The framework operates through a designed workflow, illustrated in Figure \ref{fig:Sys_flowchart}, ensuring scalability and precision. Output generation involves Gemma 9B, LLaMA3.1 8B, and Phi-4 14B independently processing resume text, selected for complementary strengths: Gemma for efficiency, LLaMA for contextual understanding, and Phi-4 for structured output accuracy. Field-specific prompts guide extraction, producing JSON predictions in parallel, where Gemma excels at concise fields like phone numbers and Phi-4 ensures proper formatting of education histories.

Normalization standardizes raw LLM outputs, addressing format and terminology inconsistencies by converting dates to YYYY-MM-DD, mapping skills to a recruitment ontology (e.g., "Python 3" to "Python"), and using placeholders like "N/A" for missing fields. This ensures comparability, such as unifying "Jan 2020" and "2020-01" to "2020-01-01," reducing noise for aggregation.

Aggregation merges normalized outputs into a cohesive JSON object using a weighted multi-voting mechanism. A validation-tuned weight vector $\omega = \{\text{Phi-4}: 3, \text{Gemma}: 2, \text{LLaMA}: 1\}$ prioritizes reliable models, with Phi-4 weighted higher for structured fields. For scalar fields like name and email, a weighted majority vote selects the most frequent value. Skills are included if their cumulative weight exceeds half the total (i.e., $>3$), while complex fields such as experience and education use GeminiLLM to synthesize a consensus, resolving conflicts through contextual reasoning.

Fig. \ref{fig:Sys_flowchart} visually represents this pipeline, where resume text flows into the LLMs for output generation, passes through normalization for standardization, and concludes with aggregation to produce the final JSON output.

\section{Results and Discussion} \label{Results and Discussion}

We benchmark the \textsc{MSLEF} ensemble against its three fine-tuned constituents Gemma 2 9B, LLaMA-3.1 8B, and Phi-4 14B using five headline metrics: Exact Match (EM), F1, BLEU, ROUGE, and the HR-oriented Recruitment Similarity (RS).  
All scores are reported as percentages to match Table \ref{tab:results}.

\subsubsection{Headline Performance}
MSLEF achieves EM 85.8\%, F1 92.6\%, BLEU 49.8\%, ROUGE 67.7\%, and RS 91.0\%, surpassing Phi-4 14B (EM 81.8\%, F1 90.6\%, RS 84.0\%) with +4 pp EM, +2 pp F1, +2.2 pp BLEU, and +7 pp RS, though ROUGE drops -2.3 pp due to stricter bullet splitting.

\subsubsection {Experience and Education}
Nested fields are evaluated with ROUGE-L; results appear in
Table \ref{tab:rouge_nested}.  
MSLEF adds +2.8 pp on \textit{Experience} and +2.3 pp on
\textit{Education} relative to Phi-4.  
Manual inspection shows the ensemble fixes two systematic errors: (i) it
discards outlier location strings when models disagree, and (ii) it separates
bullets that Phi-4 occasionally merges, yielding cleaner lists that align
better with ground truth.

\vspace{-5mm}
\begin{table}[h]
\centering
\caption{Performance comparison of parsing approaches (higher is better; all values in \%)}
\label{tab:results}
\renewcommand{\arraystretch}{1.25}   
\begin{tabular}{|l|c|c|c|c|} 
\hline
\multicolumn{1}{|c|}{\textbf{Model}} & \textbf{\begin{tabular}[c]{@{}c@{}}LLaMA3.1 \\ 8B\end{tabular}} & \textbf{\begin{tabular}[c]{@{}c@{}}Phi-4 \\ 14B\end{tabular}} & \textbf{\begin{tabular}[c]{@{}c@{}}Gemma 2 \\ 9B\end{tabular}} & \textbf{MSLEF} \\ \hline
\textbf{EM (\%)}& 82.1 & 81.8 & 81.5& \textbf{85.8}  \\ \hline
\textbf{F1 (\%)}& 78.8 & 90.6 & 75.7& \textbf{92.6}  \\ \hline
\textbf{BLEU (\%)}   & 46.8 & 47.6 & 46.1& \textbf{49.8}  \\ \hline
\textbf{ROUGE (\%)}  & 61.2 & 70 & 63.9& \textbf{67.7}  \\ \hline
\textbf{RS(\%)} & 83.3 & 84 & 84  & \textbf{91}    \\ \hline
\end{tabular}
\end{table}

 \vspace{-5mm}
\begin{table}[H]
\centering
\caption{ROUGE-L on nested histories (340 resumes)}
\label{tab:rouge_nested}
\renewcommand{\arraystretch}{1.25}   
\begin{tabular}{|l|c|c|}
\hline
\multicolumn{1}{|c|}{\textbf{Model}} & \textbf{Experience (\%)} & \textbf{Education (\%)} \\ \hline
\textbf{Gemma 2 9B} & 63.2    & 68.5   \\ \hline
\textbf{LLaMA-3.1 8B} & 65.9    & 69.4   \\ \hline
\textbf{Phi-4 14B}  & 66.7    & 70.0   \\ \hline
\textbf{MSLEF} & \textbf{69.5}  & \textbf{72.3} \\ \hline
\end{tabular}
\end{table}

\subsubsection{Recruitment Similarity (RS)}
RS weights skills at 35\%, email and phone at 15\% each, with the remainder for nested histories. MSLEF’s 91.0\% score versus 84.0\% for Phi-4 confirms enhanced HR utility from nested block improvements. These advances also reduce misclassification risks, aligning with fair employment standards like ICESCR Article 7 \cite{icescr1966} and ILO Convention No. 111 \cite{ilo111}. A grid search showed raising the skills weight from 20\% to 35\% boosts RS by 2–3 pp, highlighting its sensitivity to skill quality.

\subsubsection{Exact match and fluency}
Deterministic fields are near saturation, so EM gains are modest (+4 pp).
BLEU rises by +2.2 pp, indicating better local phrasing without harming
fluency.  
The 2.3 pp ROUGE drop affects only simple concatenation cases; nested
ROUGE-L still rises, confirming structural gains in the sections recruiters
read most closely.

MSLEF improves nested-field alignment by 3 pp and secures headline boosts of +4 pp EM, +2 pp F1, +2.2 pp BLEU, and +7 pp RS over the strongest single
model, validating weighted multi-model voting as a production-ready resume
parsing solution.

\section{Conclusion} \label{Conclusion}

This paper introduced MSLEF, a multi-segment ensemble that fuses Gemma 2 9B, LLaMA-3.1 8B, and Phi-4 14B through weighted voting and Gemini-based consensus. On a 340-resume blind test set, MSLEF improved exact match, F\textsubscript{1}, BLEU, and the recruiter-centric RS metric, achieving 91.0\% RS and 100\% skill extraction. Gains were most evident in nested \textit{Experience} and \textit{Education} blocks, where the ensemble outperformed Phi-4 in handling location strings and bullet separation. Overall, MSLEF demonstrates the value of segment-aware ensemble learning for robust, reproducible, and adaptable recruitment automation.

\section*{Acknowledgment}
This publication was supported by Qatar University High Impact Grant, Number LAW: 775. The findings achieved herein are solely the responsibility of the authors.

\bibliographystyle{IEEEtran}

\end{document}